\documentclass[letterpaper]{article} % DO NOT CHANGE THIS
\usepackage{aaai25}  % DO NOT CHANGE THIS
\usepackage{times}  % DO NOT CHANGE THIS
\usepackage{helvet}  % DO NOT CHANGE THIS
\usepackage{courier}  % DO NOT CHANGE THIS
\usepackage[hyphens]{url}  % DO NOT CHANGE THIS
\usepackage{graphicx} % DO NOT CHANGE THIS
\usepackage{natbib}  % DO NOT CHANGE THIS AND DO NOT ADD ANY OPTIONS TO IT
\usepackage{caption} % DO NOT CHANGE THIS AND DO NOT ADD ANY OPTIONS TO IT
\usepackage{algorithm}
\usepackage{algorithmic}
\usepackage{amsmath,amssymb,amsfonts}
\usepackage{algorithmic}
\usepackage{cleveref}
\usepackage{textcomp}
\usepackage{xcolor}
\usepackage{threeparttable}
\usepackage{booktabs}
\usepackage{multirow} 
\usepackage{tabularx}
\usepackage{graphicx} 
\usepackage{subcaption}
\usepackage{newfloat}
\usepackage{listings}
\DeclareCaptionStyle{ruled}{labelfont=normalfont,labelsep=colon,strut=off} % DO NOT CHANGE THIS
\floatstyle{ruled}
\newfloat{listing}{tb}{lst}{}
\floatname{listing}{Listing}
\newcommand{\cnameref}[1]{\ref{#1}}
\usepackage{amsthm}
\newtheorem{definition}{Definition}

\title{Global Confidence Degree Based Graph Neural Network for Financial Fraud Detection  }%\thanks{Sponsered by Tongji University innovation funds.}}
\author{
    Jiaxun Liu\equalcontrib, 
    Yue Tian\equalcontrib,
    Guanjun Liu \thanks{Corresponding Author is Guanjun Liu.}
}
\affiliations{
    The Department of Computer Science, Tongji University, Shanghai 201804, China \\\{ljx0316, 1810861, liuguanjun\}@tongji.edu.cn
}

\usepackage{bibentry}
\begin{document}

\maketitle

\begin{abstract}
Graph Neural Networks (GNNs) are widely used in financial fraud detection due to their excellent ability on handling graph-structured financial data and modeling multilayer connections by aggregating information of neighbors. However, these GNN-based methods focus on extracting neighbor-level information but neglect a global perspective. This paper presents the concept and calculation formula of Global Confidence Degree (GCD) and thus designs GCD-based GNN (GCD-GNN) that can address the challenges of camouflage in fraudulent activities and thus can capture more global information. To obtain a precise GCD for each node, we use a multilayer perceptron to transform features and then the new features and the corresponding prototype are used to eliminate unnecessary information. The GCD of a node evaluates the typicality of the node and thus we can leverage GCD to generate attention values for message aggregation. This process is carried out through both the original GCD and its inverse, allowing us to capture both the typical neighbors with high GCD and the atypical ones with low GCD. Extensive experiments on two public datasets demonstrate that GCD-GNN outperforms state-of-the-art baselines, highlighting the effectiveness of GCD. We also design a lightweight GCD-GNN (GCD-GNN$_{light}$) that also outperforms the baselines but is slightly weaker than GCD-GNN on fraud detection performance.  However, GCD-GNN$_{light}$ obviously outperforms GCD-GNN on convergence and inference speed.

% Finally, the model leverages information from typical (high GCD) and atypical (low GCD) neighbors and aggregates it through the transformation matrix generated by the nodes' intrinsic features, thereby improving its ability to discriminate between fraudulent and benign nodes. 

% A light version model is also proposed, which is tremendously faster and has a slight loss in performance. Extensive experiments on public datasets demonstrate that GCD-GNN outperforms state-of-the-art baselines.
\end{abstract}


\section{Introduction}
% Fraud detection is extensively applied in e-commerce\citep{cheng2020spatio, lin2021online, xu2021towards} social media\citep{kaghazgaran2017behavioral,dhawan2019spotting}, network security\citep{dou2020robust}, and other domains. Numerous methods have been proposed in this area. Among them, graph-based approaches, particularly Graph Neural Network (GNN) methods, have garnered significant attention due to their ability to capture valuable information from nodes and their neighbors.

% Financial fraud is prevalent and damaging both at the company level and personally. Economic scholars estimate that 14.5\% of large U.S. public companies engage in financial fraud, resulting in approximately a 3\% loss in enterprise value \cite{reurink2018financial}. Major firms such as Enron in 2001, Wirecard in 2019, and Evergrande in 2021 have suffered significantly from these scandals. On a personal level, the rapid increase in transaction volumes due to various payment methods complicates oversight \cite{weng2018online}. Thus, financial fraud detection is necessary to avoid the loss. 
Financial fraud is widespread and damaging, affecting both organizations and individuals. Economic scholars estimate that approximately 14.5\% of large U.S. public companies engage in financial fraud, leading to an estimated 3\% loss in enterprise value \cite{reurink2018financial}. Large-scale corporations, including Enron in 2001, Wirecard in 2019, and Evergrande in 2021, have faced significant consequences due to these scandals. On a personal level, the increasing transaction frequency associated with various payment methods complicates oversight \cite{weng2018online}. Therefore, detecting financial fraud is crucial to preventing substantial losses.

%Graph Neural Networks (GNNs) act as popular approaches for mining structural data in financial fraud detection. Unfortunately, vanilla GNNs often underperform due to the characteristics of the financial fraud activities, which include: 1) \textbf{Complex relationships}: It is challenging to directly identify the relationships between entities based solely on their connections. 2) \textbf{Camouflage activities} \cite{dou2020enhancing}: Fraudsters use strategies to obscure their fraudulent activities, making detection more difficult. To address these challenges, some improved GNN models employ attention mechanisms to assess the importance of edges during the message-passing process \citep{wang2019semi, liu2021intention}.  Others focus on enhancing homophilous connections and reducing heterophilous ones \citep{dou2020enhancing, liu2021pick}.  In addition, some models leverage label information to handle nodes in different classes separately \citep{wang2023label, zhuo_partitioning_2024}. All these works analyze financial fraud detection from node and its neighbors. There are few methods that address this issue from global scale. Nevertheless, global scale could provide sufficient information, which is proved in unsupervised anomaly detection\citep{roy_gad-nr_2024, ding2019deep}. %However, those models primarily focus on filtering aggregated messages, which proves insufficient for comprehensively understanding financial fraud activities.
GNNs are widely used for mining structural data in financial fraud detection. Traditional GNNs often underperform due to the inherent characteristics of financial fraud activities, which include complex relationships and camouflage activities. 1) \textbf{Complex relationships} \cite{ma2021comprehensive}: It is challenging to directly identify the relationships between entities based solely on their connections. 2) \textbf{Camouflage activities} \cite{dou2020enhancing}: Fraudsters employ strategies to obscure their fraudulent activities, thereby complicating detection. To address these challenges, some advanced GNN models employ attention mechanisms to assess the significance of edges during the message-passing process \citep{wang2019semi, liu2021intention}. Other models focus on enhancing homophilous connections while reducing heterophilous ones \citep{dou2020enhancing, liu2021pick}. In addition, some models use the label information to handle nodes in different categories separately \citep{wang2023label, zhuo_partitioning_2024}. All of these studies analyze financial fraud detection at the level of individual nodes and their neighbors. However, these methods overlook that neighbor messages can be harmful due to not only heterophily but also deceptive features, such as a fraudulent node camouflaged with normal features. This issue can be addressed on a global scale by evaluating the typicality of each node and eliminating messages based on their typicality, which aids in accurate classification, as demonstrated in unsupervised anomaly detection \citep{roy_gad-nr_2024, ding2019deep}.

To fill the above gap, our paper aims to address financial fraud detection from global scale. Inspired by \citep{gao_alleviating_2023,shi_h2-fdetector_2022}, we use the \textbf{prototype} to represent the global feature of a graph as global information. In our task, we seperately define two prototypes that are generated from all nodes in the same category. Following this, the article can address the following challenges: \textbf{1) How to generate an appropriate prototype to represent all nodes in a graph? } The prototype should encapsulate the maximum amount of information from nodes within the same category, with each node contributing appropriately to its corresponding prototype. Moreover, unnecessary information should be eliminated to avoid overfitting. \textbf{2) How to extract Global Confidence Degree (GCD) for each node in a graph?} We define the similarity between the prototype and each node as GCD to represent the typicality of a node. For labeled nodes, we can directly compare them with the prototype in the same category. For unlabeled nodes, we experiment with several methods to generate GCD and identify the approach that offers high performance and low time complexity. \textbf{3) How to utilize GCD in message generation?}  It is natural to maximize the extraction of the most typical information. However, atypical nodes (e.g., a node with features significantly different from its prototype) also provide valuable information.

\begin{figure}[tbp]
  \centering
  \includegraphics[width=\linewidth]{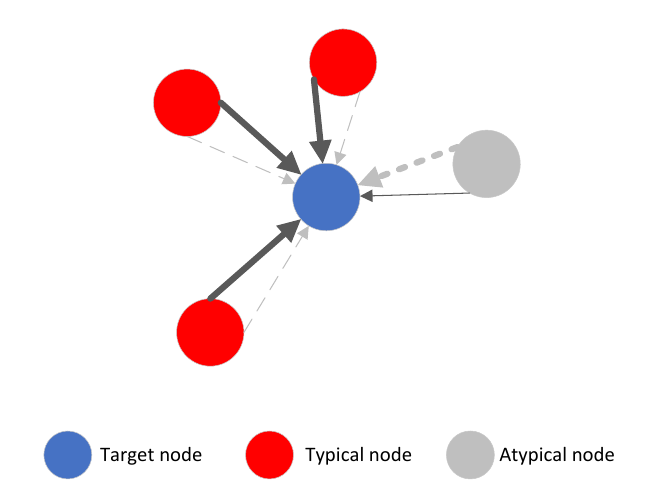} %1.png是图片文件的相对路径
  \caption{Aggregation pattern. Aggregate from typical and atypical perspectives. Solid lines represent the aggregation of the typical perspective, while dashed lines represent the aggregation of the atypical perspective. The thickness of the line is directly proportional to the weight value of a node.} %caption是图片的标题
  \label{fig:pattern} %此处的label相当于一个图片的专属标志，目的是方便上下文的引用
\end{figure}

\begin{figure}[t]
    \centering
    % \begin{subfigure}[b]
        \begin{subfigure}[b]{0.45\linewidth}
            \centering
            \includegraphics[width=\linewidth]{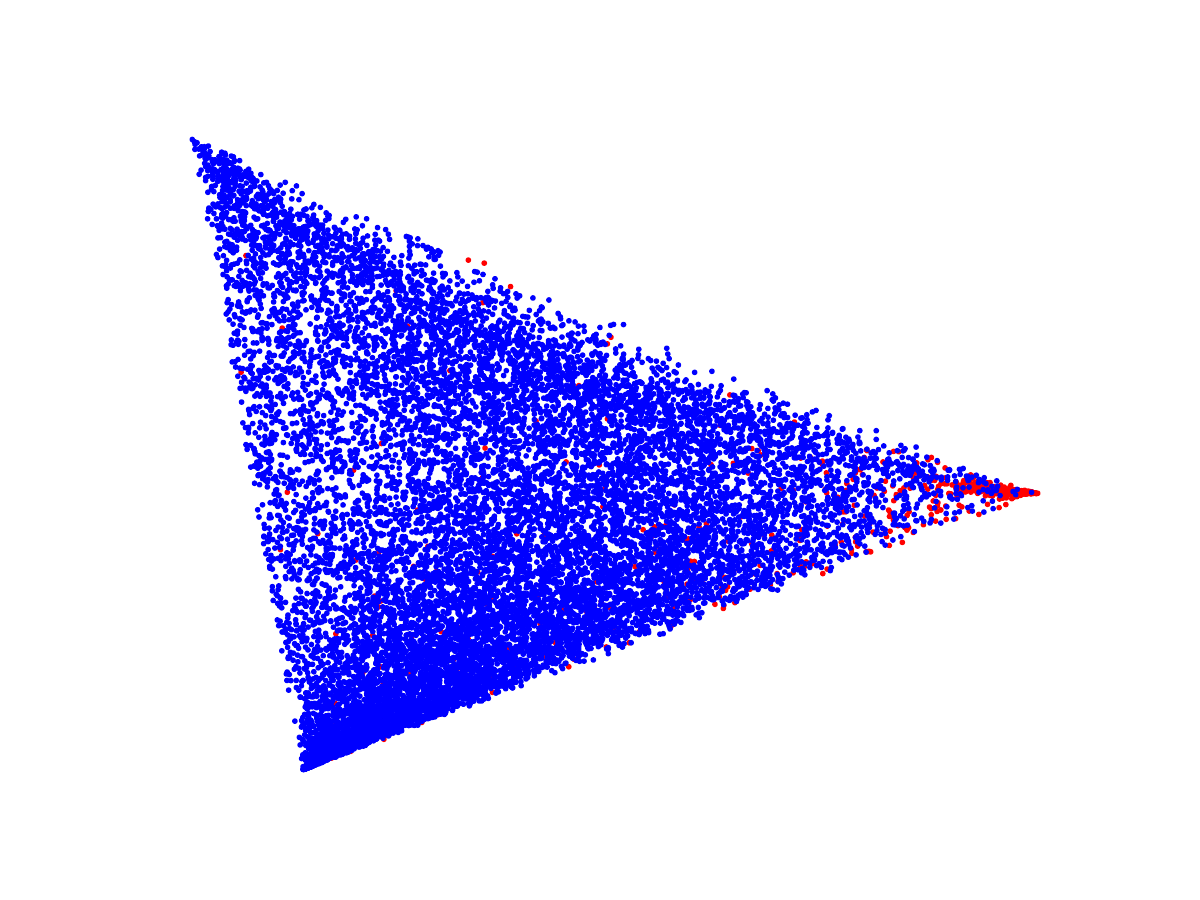}
            \caption{PCA original feature}
            % \label{}
        \end{subfigure}
        \hfill
        \begin{subfigure}[b]{0.45\linewidth}
            \centering
            \includegraphics[width=\linewidth]{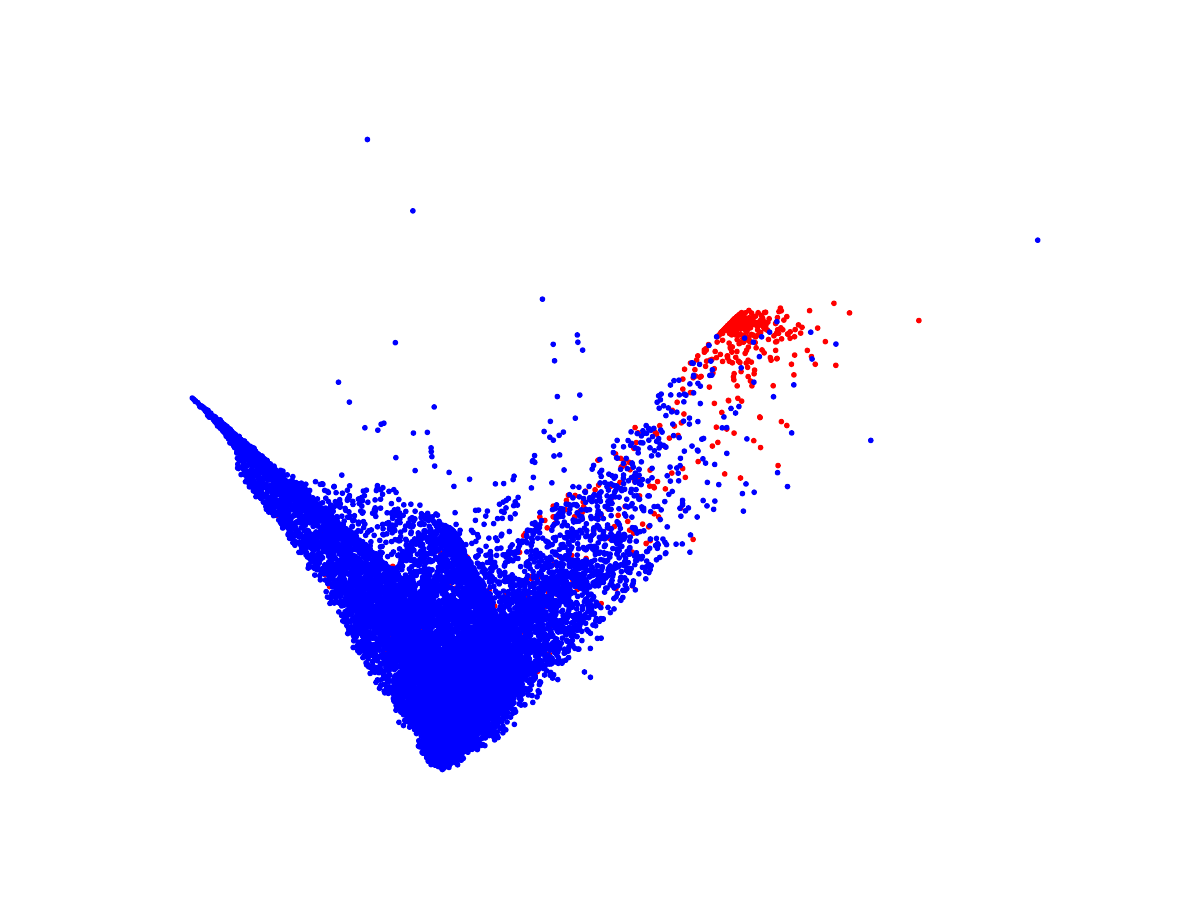}
            \caption{PCA mixed feature}
            % \label{PCA mixed feature}
        \end{subfigure}
    
        \vspace{0.0cm} % 添加垂直间距
        % \hfill
    
        \begin{subfigure}[b]{0.45\linewidth}
            \centering
            \includegraphics[width=\linewidth]{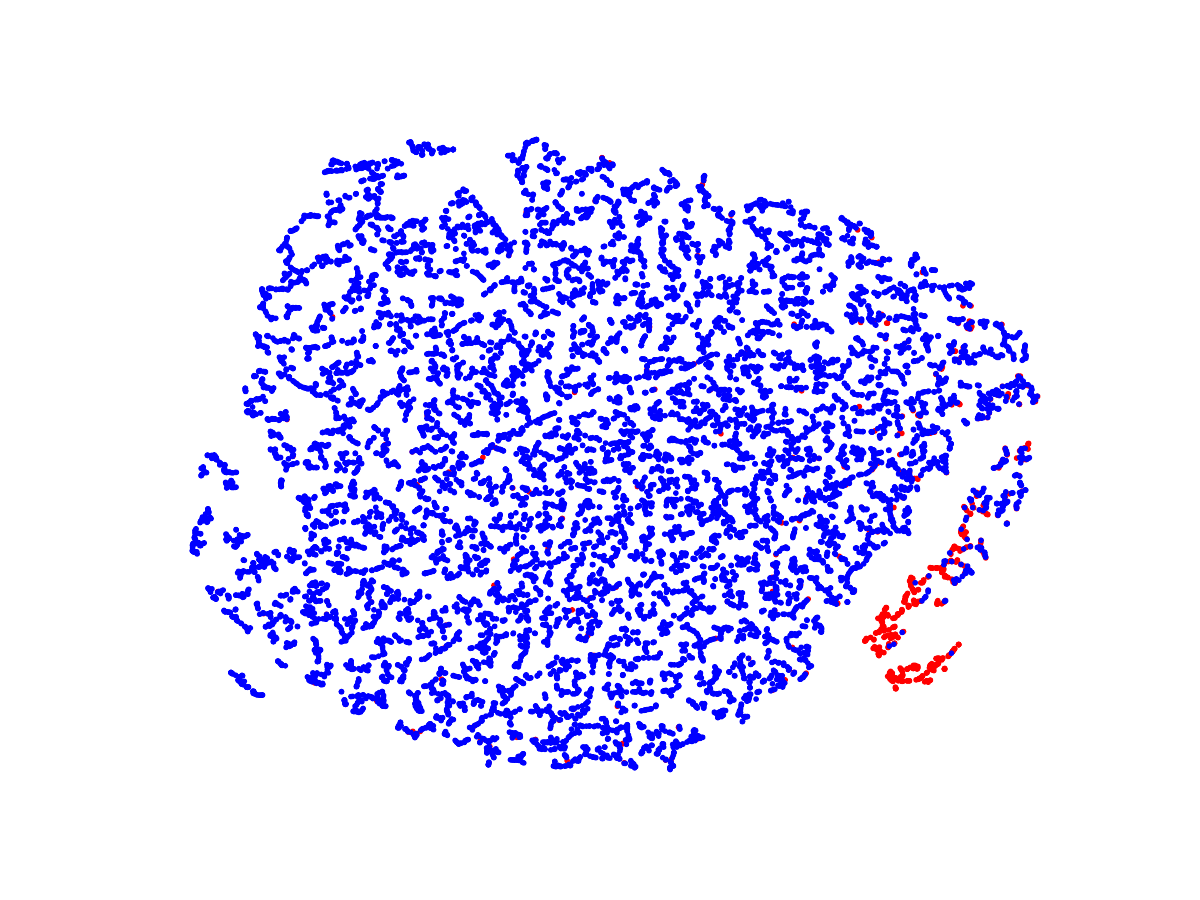}
            \caption{T-SNE original feature}
            % \label{fig:image3}
        \end{subfigure}
        \hfill
        \begin{subfigure}[b]{0.45\linewidth}
            \centering
            \includegraphics[width=\linewidth]{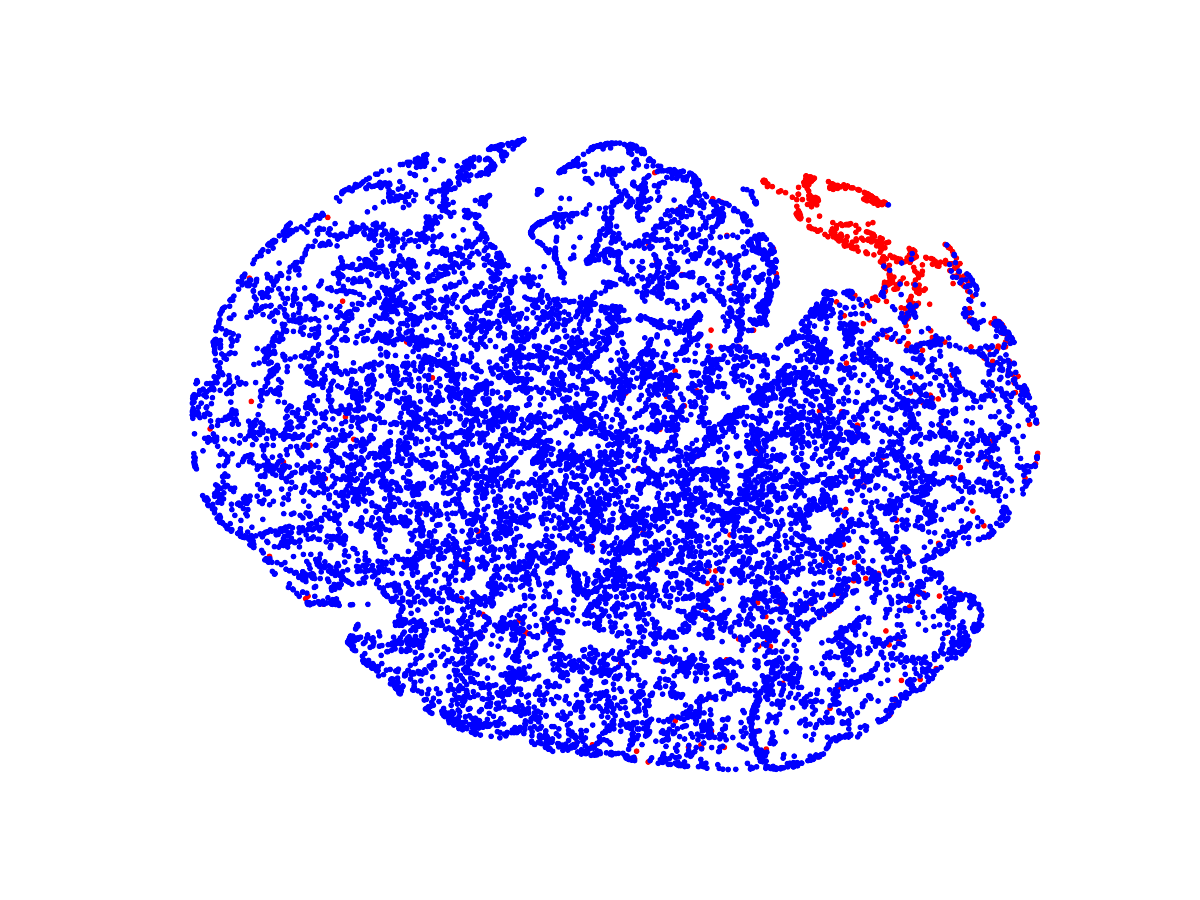}
            \caption{T-SNE mixed feature}
            % \label{fig:image4}
        \end{subfigure}
    
        \caption{Feature embeddings on T-Finance visualization using two different dimensionality reduction techniques. Red color represents fraudulent nodes, while blue represents benign nodes.}
        \label{fig:featureusage}
    % \end{subfigure}
    
\end{figure}

To tackle the above issues, we propose a \textbf{G}lobal \textbf{C}onfidence \textbf{D}egree Based \textbf{G}raph \textbf{N}eural \textbf{N}etwork \textbf{(GCD-GNN)}. Firstly, we project the original features into a new space for  extracting prototypes from these features which are then combined with the original features to be used for classification. Secondly, we propose a comparasion module to generate the GCD of each node. Thirdly, we utilize GCD to calculate weight values for aggregation. In order to utilize both typical and atypical information, We aggregate messages from the typical and atypical perspectives separately %and combine them through a matrix transformation, 
as illustrated in Fig.\ref{fig:pattern}. Inspired by \citep{chen_consistency_2024, zhuo_partitioning_2024}, we employ a transformation matrix generated from the node's intrinsic features, as a component for message aggregation, ensuring that the node's own information directly influences the aggregation process. %Therefore, the update of the target node feature is additionally influenced by the characteristics of itself.%To fully exploit the information contained in the relationship among source and destination nodes and capture local information, the transformation matrix generated by the nodes' intrinsic features, which is used for message aggregation. The generated message consider the features from all source and destination nodes.

% \onecolumn

% \twocolumn

Our main contributions are summarized as follows:

\begin{itemize} 
    \item We transform features to generate better prototypes, Those new features can also eliminate the unnecessary information and increase the separation between fraudulent nodes and benign ones. Results are visualized in Fig. \ref{fig:featureusage}. Therefore, the GNN can more effectively identify fraudulent nodes within the graph.%the GNN can more effectively identify features in different categories. 
    \item We utilize the GCD of each node to extract information on a global scale, which offers a novel perspective for observing fraud patterns, ensures model performance and significantly enhances convergence speed.
    \item We aggregate both typical and atypical information as shown in Fig. \ref{fig:pattern}, This approach enriches the message source and removes disruptive information, directly enhancing model performance.
\end{itemize}

In addition, extensive experiments are conducted on two open datasets. The outcome shows that our model outperforms the state-of-the-art model. 

To accommodate different requirements, we provide two versions of our methods. The lightweight version delivers solid performance with fast processing, while the full version provides superior performance among baseline models with relatively fast speed.

\section{Related Work}

% \subsection{Fraud Detection}
% According to \citep{abdallah2016fraud}. Fraud detection systems (FDS) face the following challenges. 1) Concept Drift: This refers to the phenomenon where the underlying model or concept changes over time due to shifts in user behavior or fraudulent tactics, necessitating adaptive learning algorithms to update detection models dynamically. 2) Skewed Class Distribution: Imbalanced class issues occur when there are significantly fewer samples of fraudulent instances compared to normal instances, leading to difficulties in training classifiers to accurately detect fraud. 3) Reduction of Large Amounts of Data: The vast scale and high dimensionality of fraud data sets can complicate the mining and detection process, requiring data reduction approaches such as dimensionality reduction and numerosity reduction. 4) Noisy Data: Fraud data sets can be noisy, incomplete, and inconsistent, necessitating data cleaning routines and smoothing techniques to improve the accuracy of FDS. 5) Support for Real-Time Detection: Some applications require immediate fraud detection and response, which demands efficient online fraud detection solutions that can operate with limited resources like time and memory.

% \subsection{Financial Fraud}

% companies dataset\cite{wu2023splitgnn}, transaction fraud detection\cite{tang2022rethinking}
%!!!!!!要改名字!!!!!!!hybrid views
\begin{figure*}[th]
  \centering
  \includegraphics[width=\linewidth]{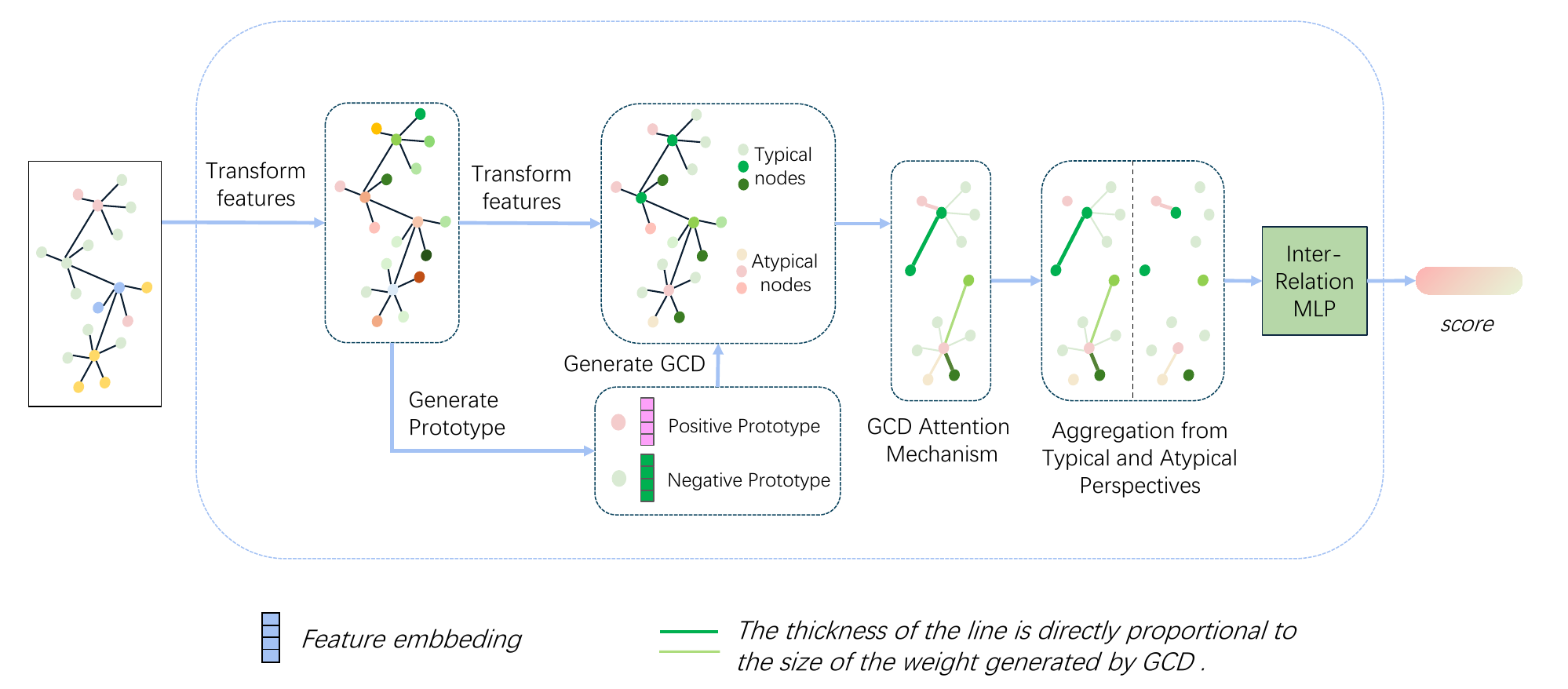} %1.png是图片文件的相对路径
  \caption{An illustration of the proposed framework.} %caption是图片的标题
  \label{modelstru} %此处的label相当于一个图片的专属标志，目的是方便上下文的引用
\end{figure*}

\subsection{Financial Fraud Detection}
Several machine learning techniques have been proposed to address the problem of financial fraud detection. For example, \cite{Zaki_Meira_2014} compare neural network-based models and decision tree models, finding that neural networks outperform decision trees. Additionally, a signature-based method for detecting potential fraud in e-commerce applications was proposed by \cite{mota2014usage}. This approach provides an alternative method for detecting fraudulent activities by identifying deviations in user behavior, thereby enabling real-time detection of potential fraudulent activities. 
Moreover, A deep learning-based model that integrates numerical financial data with textual information from management discussions \cite{xiuguo2022analysis} has been developed to enhance the detection of financial statement fraud among Chinese listed companies. This model demonstrates significant improvements over traditional methods.
Futhermore, \cite{yu_group-based_2023} presents a novel semi-supervised Group-based Fraud Detection Network (GFDN) that leverages structural, attribute, and community information from attributed bipartite graphs to effectively detect group-based financial fraud on e-commerce platforms. %statement fraud的

\subsection{Graph Anomaly Detection}
Fraudulent activities have become increasingly frequent, leading to the development of various detection methods. Rule-based and outlier detection techniques, as summarized in \citep{kou2004survey, phua2010comprehensive}, highlight models based on machine learning approaches, including support vector machines (SVM) and decision trees. 

Recently, graph neural networks(GNN) have been utilized in fraud detection. For instance, Care-GNN \cite{dou2020enhancing} and Rio-GNN \cite{10.1145/3490181} exploit reinforcement learning to detect camouflage activities within networks. PCGNN \citep{liu2021pick} connects homophilic nodes and filters out heterophilic nodes to enhance the message passing process. Additionally, \citep{wang_label_2023,zhuo_partitioning_2024} utilize label information, dividing nodes into separate groups based on their labels and separately processing messages generated from different groups. 

%about prototype
The prototype has been employed in previous studies \citep{gao_alleviating_2023, shi_h2-fdetector_2022} for feature optimization, enhancing the network's ability to distinguish between fraudulent and benign nodes. However, these methods incorporate the prototype only within the training loss, neglecting the critical confidence information that indicates whether a node in the graph is typical or atypical. This oversight restricts the potential benefits of using the prototype for more nuanced and effective differentiation.

\section{Methodology} \label{chap:met}
Previous models often encounter the issue of message elimination in resource-intensive methods like reinforcement learning or graph transformers. In contrast, some newest models avoid message elimination by dividing neighbors into distinct groups and aggregating their information separately. These operations also increase model complexity and extend training and inference time. However, by using GCD, our model achieves better performance, enabling faster training and inference simultaneously.

In this section, we outline the GCD-GNN framework. First, we define the role of GCD within the fraud detection context in Section \cnameref{chap:met:GCFr}. Then, an overview of the entire model is provided in Section \cnameref{chap:met:overview}. Finally, we detail the key components in Sections \cnameref{chap:met:extfeat}-- \cnameref{chap:met:GCAgg}.

\subsection{Prototype and Global Confidence Degree (GCD)} \label{chap:met:GCFr}
%We need a value to represent the typicality of a node feature across the entire graph. To achieve this goal, We define the prototype in the following chapters as an aggregation of information from nodes within the same category in the graph. And we define GCD by comparing the prototype and a node.

%\subsubsection{\textit{Definition}} 
\begin{definition}[Multi-relation Graph]
    We define a multi-relation graph as $\mathcal{G} = (\mathcal{V}, \mathcal{X}, \{\mathcal{E}_r\}_{r=1}^{R}, \mathcal{Y})$. $\mathcal{V}$ is the set of nodes $\{v_1, \ldots, v_n\}$. Each node $v_i$ has a $d$-dimensional feature vector $\mathbf{x}_i \in \mathbb{R}^d$ and $\mathcal{X} = \{\mathbf{x}_1, \ldots, \mathbf{x}_n\}$ is the features. $e_{i,j}^{r} = (v_i, v_j) \in \mathcal{E}_r$ is an edge between $v_i$ and $v_j$ with a relation $r \in \{1, \ldots, R\}$. Note that an edge can be associated with multiple relations and there are $R$ different types of relations. $\mathcal{Y}=\{\mathbf{y}_1,...,\mathbf{y}_n$\} is the set of labels for each node in $\mathcal{V}$. 
\end{definition}
In our scenario, $\mathcal{Y} \in \{fr, be, un\}$, where $fr$ means fraud labeled nodes, $be$ means benign labeled nodes and $un$ means unlabeled nodes.

\begin{definition}[Prototype]
    We define Prototype as $\mu\in \mathbb{R}^d$. \(\Phi\): $\mathbb{R}^d \rightarrow \mathbb{R}^d$ refers to the transformation applied to features. \(\xi\): $\mathbb{R}^{n\times d} \rightarrow \mathbb{R}^{d}$ aggregates features into a single vector. %a feature generated by the nodes in the whole graph in a specific way.
    \begin{align}
        \mu_{fr} = \xi(\text{concat}(\Phi(\mathbf{x}_{i})))\ \ \ \hfill \mathbf{y}_i=fr,\notag \\
        \mu_{be} = \xi(\text{concat}(\Phi(\mathbf{x}_{j})))\ \ \ \mathbf{y}_j=be. 
        \label{eq:proto}
    \end{align}
    
\end{definition}

Further details of \(\sigma\) and \(\xi\) are provided in Section  \cnameref{chap:met:extfeat}.
%\subsubsection{\textit{Definition}} 
\begin{definition}[Global Confidence Degree] \label{def:GCD}
    we denote Global Confidence Degree (GCD) as $g$. \(g_i\in \mathbb{R}\) is the GCD value of the i-th node in the graph. $\sigma: \mathbb{R}^{d}\times \mathbb{R}^{d} \rightarrow \mathbb{R}, $ means the similarity function that measures the difference of two features.
    \begin{align}
        g_i =
        \begin{cases}
            \sigma(\mu_{\mathbf{y}_i}, \mathbf{x}_i) \hfill \text{if } \mathbf{y}_i = fr \ \text{or} \ be, \\
            \max(\sigma(\mu_{fr}, \mathbf{x}_i), \sigma(\mu_{be}, \mathbf{x}_i))  \\ \hfill \text{if } \mathbf{y}_i = un. 
        \end{cases}
        \label{eq:gci}
    \end{align}
\end{definition}
$g_i$ represents the typicality of the node $i$. For labeled nodes, we use the similarity between each node and its corresponding prototype. For unlabeled nodes, we select the maximum between \( \sigma(\mu_{fr}, x_i) \) and \( \sigma(\mu_{be}, x_i) \). %our experiments show that the maximum similarity serves as their GCD. 
Details about the similarity function can be found in Section \cnameref{chap:met:utlGC}. 
%The details about similarity function is in Section.\cnameref{chap:met:extfeat}.
%We discuss different similarity function in \cnameref{chap:met:extfeat}. ！！！need to do

\subsection{Overview}
\label{chap:met:overview}
GCD-GNN includes a prototype calculator, a GCD estimator, a special GNN layer and a multilayer perceptron (MLP) aggregator. The prototype and GCD estimator contains an iterative prototype generator and a GCD generator depends on the similarity between nodes and their corresponding prototypes. The special GNN layer based on GraphSAGE \citep{hamilton2017inductive}, contains a message generator utilizing two kinds weight values generated by original GCD and its reverse.
An aggregator receives messages derived from two kinds of weight values. The detailed structure of our method is shown in Fig. \ref{modelstru}.

\subsection{Extracting Prototype Feature}
\label{chap:met:extfeat}
Inspired by \citep{gao_alleviating_2023}, to extract the prototype feature, we exploit the iterative extraction of the prototype. Firstly, we use an MLP and Graph Normalization \citep{cai2021graphnorm} to process the initial features, projecting those features into a space that is fitting for measuring similarity.
\begin{align}
    \mathbf{x}_{exp} = GraphNorm(\Phi(\mathcal{X})), 
\end{align}
where $\Phi$ indicates an MLP. In addition, prototypes are generated by calculating the mean value of node features for the corresponding category. After this initial state, prototypes are iteratively updated based on node similarity, as shown in Eq.\ref{eq:proto}. Here, $\Phi$ represents an MLP. For the initial state, $\xi$ employs the Mean function, which calculates the average value of a set of features. For subsequent updates, we adopt the strategy proposed by \cite{gao_alleviating_2023}, with the following specifics:

\begin{align}
    \label{eq:iteproto}
    s_v^{(e)}=cos(\mathbf{x}_v^{(e)},\mu^{(e-1)}), \notag \\
    w_v^{(e)}=\frac{exp(s_v^{(e)}/\tau)}{\sum_{u=1}^N exp(s_u^{(e)/\tau})}, \notag \\
    \mu^{(e)}=\sum_{v=1}^{N} w_v\cdot \mathbf{x}_v^{(e)},
\end{align}
where \(\tau\) is the temperature parameter that controls the smoothness of the weights. First, we compute the cosine similarity between each node \(v\) and the previous prototype \(\mu^{(e-1)}\), as shown in the first part of Eq. \ref{eq:iteproto}. The softmax output of this similarity serves as the weight \(w_v^{(e)}\) for each node. Finally, the updated prototype \(\mu^{(e)}\) is calculated as a weighted sum of the node features.

\subsection{Utilizing the Projected Node Feature} \label{chap:met:utlGCfeat}
We propose a weight mix method to leverage the projected features while preserving the essential characteristics of the original features that might be lost during projection. The details of this method are as follows:
\begin{align}
    \lambda=Sigmoid(\Phi(\mathbf{x})), \notag\\
    \mathbf{x}_{mixed} = \lambda \cdot \mathbf{x}_{exp}+ (1-\lambda) \cdot \mathbf{x}, 
\end{align}
where $\Phi$ indicates an MLP which generates a number that is processed by a $Sigmoid$ function to range from 0 to 1 and this result determine the weight $\lambda$. Subsequently, we combine the two features by adding them according to the weight $\lambda$. We use PCA and T-SNE to visualize the effect of the mixed feature. The details are in Fig. \ref{fig:featureusage}.

\subsection{Global Confidence Degree Calculation} \label{chap:met:utlGC}
To calculate the GCD, we need to calculate similarity by comparing each node feature with the corresponding prototype. We calculate the similarity value using the cosine function as follows: 

\begin{align}
    \sigma(\mu_c, \mathbf{x}_i) = \cos(\mu_c, \mathbf{x}_i).
\end{align}
% \begin{align}
%     simi(proto_c, x_i) = Softmax(LeckyRelu(simi(proto_c, x_i)))
% \end{align}

 The strategy for processing labeled and unlabeled data to calculate GCD is mentioned in Eq. \ref{eq:gci}.

 %说明一下lightversion
%Our lightweight model is formed with all components mentioned previously. The lightweight model has fast training and inference speed and could achieve solid performance, the details of which are in Sections \cnameref{chap:exp:perComp}, \cnameref{chap:exp:abla}.

\subsection{Aggregation from Typical and Atypical Perspectives} \label{chap:met:GCAgg}
To utilize GCD, two perspectives, termed typical and atypical, are employed for message generation. In the typical perspective, GCD is unchanged from the original one defined in Def. \ref{def:GCD}. The atypical GCD is the inversion of the typical GCD, i.e. represented as the negative of the original GCD.

\begin{align}
    g_i^{typ} = g_i, \notag \\ g_i^{atyp} = -g_i.
\end{align}

When a node needs to aggregate messages, the GCD of its neighbors is used to generate the corresponding message weights. In order to make the weights generated by the GCD more effective, according to \cite{velivckovic2017graph}, we use a GCD attention mechanism similar to graph attention network. %\(LeckyRelu\) activation function and \(Softmax\) function are exploited for weight generating.

\begin{align}
    \label{eq:attn}
    w_{ij} = LeckyRelu(g_j), \notag \\ \alpha_{ij} =  \frac{exp(w_{ij})}{\sum_{k\in \mathcal{N}_i} exp(w_{ik})}, 
\end{align}
where \(i\) is a target node and $j$ is one of its neighbors. \(\mathcal{N}_i\) means the neighbor set of node \(i\). $\alpha_{ij}$ means the final weight used in message aggregation. When we use $g_i^{typ}$ in the Eq. \ref{eq:attn}, we denote the weight as $\alpha_{ij}^{typ}$. Similarly, $g_i^{atyp}$ corresponds to $\alpha_{ij}^{atyp}$.

To utilize local information, according to \citep{zhuo_partitioning_2024}, a self-feature matrix is calculated by multiplying the node feature by the trained parameter. The message passing period is affected by the node feature.

\begin{align}
    \label{eq:selfMatrixGeneration}
    % used for self root
    W_i^{typ}=\Psi_i^{typ}(\mathbf{x}_i),\notag \\ W_i^{atyp}=\Psi_i^{atyp}(\mathbf{x}_i),
\end{align}
where $\Psi_i^{\text{typ}}(\mathbf{x}_i)$ and $\Psi_i^{\text{atyp}}(\mathbf{x}_i)$: $\mathbb{R}^d \rightarrow \mathbb{R}^{d \times d'}$ are two learnable weight generators. Each node receives an individual transformation weight matrix.

The message generation process, which utilizes both typical and atypical perspectives, is as follows: 
%先生成消息在相乘
\begin{align}
    m_i = W_i^{typ}\sum_{j\in\mathcal{N}_i}(\alpha _{ij}^{typ} \mathbf{x}_j) +     W_i^{atyp}\sum_{j\in\mathcal{N}_i}(\alpha _{ij}^{atyp} \mathbf{x}_j). 
\end{align}

\subsection{Lightweight Model}
The lightweight version of our method consists of prototype extracting, feature optimization and GCD attention machenism, mentioned in Section \ref{chap:met:extfeat}--\ref{chap:met:utlGC}, Building on this foundation, the full version adds self-feature matrix and aggregation from typical an atypical perspective, mentioned in \ref{chap:met:GCAgg}. The lightweight model has fast training and inference speed and could achieve solid performance. The details of which are in Sections \ref{chap:exp:perComp}, \ref{chap:exp:abla}.
 % GraphSage \citep{hamilton2017inductive} is used as the backbone of GNN, and message generation is influenced by GCD values.

% To enhance the adaption of the model to different tasks, we utilize GCD values in different ways. In financial fraud detection, non-typical nodes often represent anomalies or outliers that can introduce noise and disrupt the classification process. Eliminating these atypical nodes can enhance the accuracy and reliability of fraud detection models by focusing on more representative data patterns. Conversely, in review fraud detection, it is crucial to consider both typical and non-typical information. atypical reviews, despite being outliers, can contain valuable insights and indicators of fraudulent behavior that are not present in typical reviews. Therefore, including both types of information can provide a more comprehensive understanding of the underlying patterns of review fraud, leading to more effective detection strategies.

\section{Experiment}  \label{chap:exp}

%!!!分析数据集的特征，为什么关系模式难以提取
\subsection{Experimental Setup} \label{chap:exp:expsetup}
% YelpChi \citep{rayana_collective_2015}, Amazon \citep{mcauley2013amateurs} T-finance\citep{tang2022rethinking}, FDCompCN\citep{wu2023splitgnn}
% The YelpChi dataset \citep{rayana_collective_2015} aims to identify anomalous reviews that unjustly promote or demote certain products or businesses on Yelp.com. This dataset includes three edge types: R-U-R (reviews posted by the same user), R-S-R (reviews of the same product with the same star rating), and R-T-R (reviews of the same product posted in the same month).

% The Amazon dataset \citep{mcauley2013amateurs} focuses on detecting anomalous users who are paid to write fake product reviews in the Musical Instrument category on Amazon.com. This dataset comprises three relations: U-P-U (users reviewing at least one same product), U-S-U (users with at least one same star rating within one week), and U-V-U (users with top-5\% mutual review similarities).
\subsubsection{Datasets}
\begin{itemize}
    \item \textbf{T-Finance dataset} \citep{tang2022rethinking} aims to identify anomalous accounts in transaction networks. The nodes represent unique anonymized accounts, each characterized by 10-dimensional features related to registration days, logging activities, and interaction frequency. The edges in the graph denote transaction records between accounts. Human experts annotate nodes as anomalies if they fall into categories such as fraud, money laundering, or online gambling.

    \item \textbf{FDCompCN dataset} \citep{wu2023splitgnn} detect financial statement fraud in Chinese companies. This dataset constructs a multi-relation graph based on supplier, customer, shareholder, and financial information from the China Stock Market and Accounting Research (CSMAR) database. It includes data from 5,317 publicly listed companies on the Shanghai, Shenzhen, and Beijing Stock Exchanges between 2020 and 2023. FDCompCN features three relations: C-I-C (investment relationships), C-P-C (companies and their disclosed customers), and C-S-C (companies and their disclosed suppliers). 
\end{itemize}

Detailed statistics for the two datasets are presented in Appendix.

\subsubsection{Comparison Methods}
We compare our method with two types of models. (1) general models, including GCN \cite{kipf2016semi}, GAT \cite{velivckovic2017graph}, and GraphSAGE \cite{hamilton2017inductive}; and (2) those specifically optimized for fraud detection using GNNs, including Care-GNN \cite{dou2020enhancing}, PC-GNN \cite{liu2021pick}, BWGNN \cite{tang2022rethinking}, Split-GNN \cite{wu2023splitgnn}, 
GHRN \cite{gao2023addressing}, and PMP \cite{zhuo_partitioning_2024}. For detailed descriptions of these baselines, please refer to Appendix. 

According to \citep{tang2022rethinking}, we adopt data splitting ratios of 40\%:20\%:40\% for the training, validation, and test sets in the supervised scenario. To ensure consistency in our evaluations, each model underwent 5 trials with different random seeds. We present the average performance and standard deviation for each model as benchmarks for comparison. For clarity in the paper, all average values in the tables have been scaled by a factor of 100, and standard deviations by a factor of 10.

% We adopted data splitting ratios of 40\%:20\%:40\% for the training, validation, and test sets in the supervised scenario. We will use the results of a single experiment as the evaluation criterion for comparision.

\subsection{Performance Comparasion} \label{chap:exp:perComp}
%introduce every model

The details of our model are introduced in Section \cnameref{chap:met}. Two kinds of GCD-GNN are provided. The lightweight model, GCD-GNN\(_{light}\), contains feature optimization and GCD attention mechanisms. The full model, GCD-GNN, which includes all components, additionally integrates self-feature matrix and aggregation from typical and atypical perspectives on the basis of the lightweight model.

\setlength{\tabcolsep}{1mm}
\begin{table*}[ht]
    \centering
    \caption{Experiment results on T-Finance and FDCompCN.}
    \begin{tabular}{lcccccc}
        \toprule
        & \multicolumn{3}{c}{T-Finance} & \multicolumn{3}{c}{FDCompCN} \\
        \midrule
        Method & AUC & F1-Macro & G-Mean & AUC & F1-Macro & G-Mean \\
        \midrule
        GCN & 92.76$_{\scriptsize{\pm 0.13}}$ & 65.63$_{\scriptsize{\pm 1.15}}$ & 84.28$_{\scriptsize{\pm 0.27}}$ & 59.60$_{\scriptsize{\pm 0.27}}$ & 45.84$_{\scriptsize{\pm 0.49}}$ & 56.67$_{\scriptsize{\pm 0.24}}$ \\
        GAT & 93.04$_{\scriptsize{\pm 0.28}}$ & 77.70$_{\scriptsize{\pm 0.50}}$ & 83.52$_{\scriptsize{\pm 1.00}}$ & 59.08$_{\scriptsize{\pm 0.19}}$ & 45.97$_{\scriptsize{\pm 0.47}}$ & 52.66$_{\scriptsize{\pm 0.30}}$\\
        GraphSAGE & 84.02$_{\scriptsize{\pm 0.33}}$ & 70.56$_{\scriptsize{\pm 0.90}}$ & 79.67$_{\scriptsize{\pm 0.53}}$ & 63.31$_{\scriptsize{\pm 0.09}}$ & 45.97$_{\scriptsize{\pm 0.26}}$ & 52.66$_{\scriptsize{\pm 0.30}}$ \\
        \midrule
        Care-GNN & 87.22$_{\scriptsize{\pm 0.51}}$ & 74.42$_{\scriptsize{\pm 0.72}}$ & 60.71$_{\scriptsize{\pm 1.31}}$ & 57.36$_{\scriptsize{\pm 0.05}}$ & 47.79$_{\scriptsize{\pm 0.15}}$ & 50.96$_{\scriptsize{\pm 0.39}}$ \\
        PC-GNN & 93.49$_{\scriptsize{\pm 0.07}}$ & 81.57$_{\scriptsize{\pm 0.38}}$ & 80.97$_{\scriptsize{\pm 0.73}}$ & 59.76$_{\scriptsize{\pm 0.58}}$ & 23.83$_{\scriptsize{\pm 0.92}}$ & 54.69$_{\scriptsize{\pm 0.53}}$ \\
        BWGNN & 93.68$_{\scriptsize{\pm 0.15}}$ & 84.15$_{\scriptsize{\pm 0.31}}$ & 78.79$_{\scriptsize{\pm 0.51}}$ & 61.59$_{\scriptsize{\pm 0.62}}$ & 44.88$_{\scriptsize{\pm 1.18}}$ & 54.69$_{\scriptsize{\pm 0.53}}$ \\
        Split-GNN & 95.51$_{\scriptsize{\pm 0.07}}$ & 82.29$_{\scriptsize{\pm 0.05}}$ & 84.47$_{\scriptsize{\pm 0.25}}$ & 62.85$_{\scriptsize{\pm 0.39}}$ & 45.40$_{\scriptsize{\pm 0.57}}$ & 55.56$_{\scriptsize{\pm 0.70}}$ \\
        GHRN & 95.78$_{\scriptsize{\pm 0.08}}$ & 89.01$_{\scriptsize{\pm 0.03}}$ & 84.86$_{\scriptsize{\pm 0.11}}$ & 62.09$_{\scriptsize{\pm 0.57}}$ & 47.45$_{\scriptsize{\pm 0.85}}$ & 54.60$_{\scriptsize{\pm 0.48}}$ \\
        PMP & 97.07$_{\scriptsize{\pm 0.01}}$ & 91.96$_{\scriptsize{\pm 0.04}}$ & 88.53$_{\scriptsize{\pm 0.09}}$ & 54.34$_{\scriptsize{\pm 0.06}}$ & 48.38$_{\scriptsize{\pm 0.14}}$ & 12.02$_{\scriptsize{\pm 1.05}}$ \\
        GCD-GNN$_{light}$ (Ours) & 97.06$_{\scriptsize{\pm 0.01}}$ & 92.13$_{\scriptsize{\pm 0.01}}$ & 88.45$_{\scriptsize{\pm 0.07}}$ & 71.01$_{\scriptsize{\pm 0.12}}$ & 58.12$_{\scriptsize{\pm 0.15}}$ & \textbf{62.51$_{\scriptsize{\pm 0.31}}$} \\
        GCD-GNN (Ours)& \textbf{97.26$_{\scriptsize{\pm 0.01}}$} & \textbf{92.37$_{\scriptsize{\pm 0.05}}$} & \textbf{88.62$_{\scriptsize{\pm 0.11}}$} & \textbf{71.72$_{\scriptsize{\pm 0.18}}$} & \textbf{59.68$_{\scriptsize{\pm 0.31}}$} & 57.99$_{\scriptsize{\pm 0.31}}$ \\
        \bottomrule
    \end{tabular}
    \label{table:restf}
\end{table*}

\setlength{\tabcolsep}{2mm}
\begin{table*}[htb] 
    \centering
    \caption{Ablation results on T-Finance and FDCompCN.}
    \begin{tabular}{lcccccc}
        \toprule
        & \multicolumn{3}{c}{T-Finance} & \multicolumn{3}{c}{FDCompCN} \\
        \midrule
        Method & AUC & F1-Macro & G-Mean & AUC & F1-Macro & G-Mean \\
        \midrule
        GraphSAGE & 84.02$_{\scriptsize{\pm 0.33}}$ & 70.56$_{\scriptsize{\pm 0.90}}$ & 79.67$_{\scriptsize{\pm 0.53}}$ & 63.31$_{\scriptsize{\pm 0.09}}$ & 45.97$_{\scriptsize{\pm 0.26}}$ & 52.66$_{\scriptsize{\pm 0.30}}$ \\
        \midrule
        M1 & 97.06$_{\scriptsize{\pm 0.01}}$ & 92.13$_{\scriptsize{\pm 0.01}}$ & 88.45$_{\scriptsize{\pm 0.07}}$ & 71.01$_{\scriptsize{\pm 0.12}}$ & 58.12$_{\scriptsize{\pm 0.15}}$ & \textbf{62.51$_{\scriptsize{\pm 0.31}}$} \\
        M2 & 97.14$_{\scriptsize{\pm 0.01}}$ & 92.07$_{\scriptsize{\pm 0.03}}$ & 88.19$_{\scriptsize{\pm 0.10}}$ & 70.58$_{\scriptsize{\pm 0.28}}$ & 58.86$_{\scriptsize{\pm 0.26}}$ & 58.48$_{\scriptsize{\pm 0.44}}$ \\
        M3 & \textbf{97.26$_{\scriptsize{\pm 0.01}}$} & \textbf{92.37$_{\scriptsize{\pm 0.05}}$} & \textbf{88.62$_{\scriptsize{\pm 0.11}}$} & \textbf{71.72$_{\scriptsize{\pm 0.09}}$} & \textbf{59.68$_{\scriptsize{\pm 0.28}}$} & 57.99$_{\scriptsize{\pm 0.22}}$ \\
        \bottomrule
    \end{tabular}
    \label{table:ablation}
    % \parbox{\linewidth}{
    
    % }
\end{table*}

The results are reported in Table \ref{table:restf}, which demonstrate that our light version model performs better than baseline models on most  metrics in the public datasets. Furthermore, our complete model comprehensively surpasses the lightweight model and outperforms the baseline models across all metrics.

\begin{figure}[htbp]
    \centering
    \includegraphics[width=0.8\linewidth]{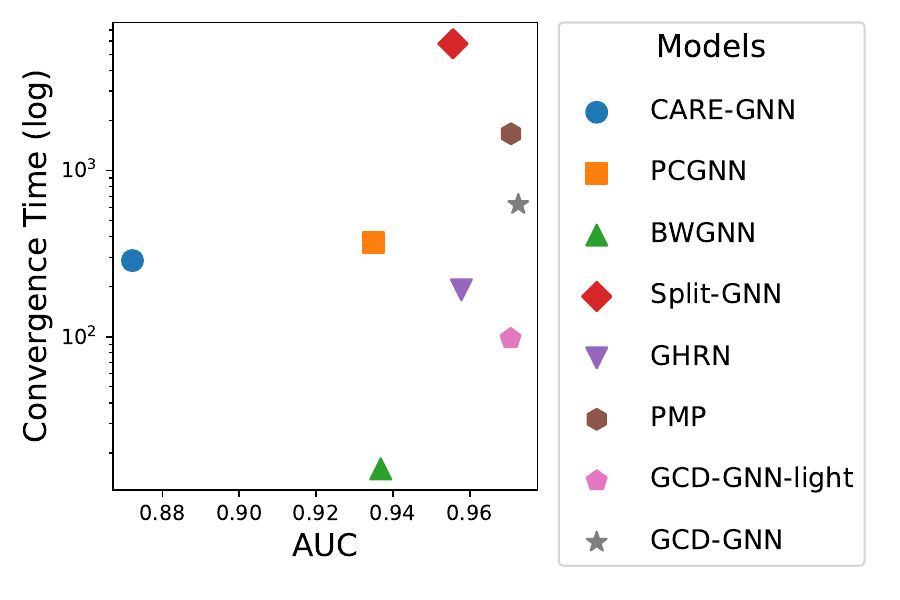}
    \caption{Convergence time (log) and AUC of models on T-Finance.}
    \label{fig:timecomsuption}
\end{figure}

We also compared the convergence speed of all models on the T-Finance dataset. The results are presented in Fig. \ref{fig:timecomsuption}, with detailed values provided in the Appendix. The results indicate that our lightweight model achieves a high AUC level in a short period of time. Furthermore, the full model achieves the highest score within a medium timeframe. The outstanding performance of our model arises from the fact that generic GNNs fail to consider the importance of each sample and aggregate messages uniformly. In contrast, our model leverages the GCD to evaluate whether the information from neighboring nodes is typical or not, which significantly improves the performance and boosts the training speed, thereby reducing computational resource consumption.

\subsection{Ablation Study} \label{chap:exp:abla}
%现有方案是用org_feat算一个feature然后融合
% We further investigate and validate three crucial components. We employ prototype extraction, feature optimization and GCD attention mechanisms, utilizing an MLP to transform features as introduced in Sections \ref{chap:met:extfeat}, \cnameref{chap:met:utlGCfeat}. Subsequently we use the new features and generated prototype to compute GCD, mentioned in Section \cnameref{chap:met:utlGC}, to assist in aggregation. Additionally, we implement self-feature matrix, using nodes' intrinsic features to generate message transformation matrices, as described in Eq. \ref{eq:selfMatrixGeneration}. Furthermore, we generate messages from both typical and atypical aspects, as described in Section \cnameref{chap:met:GCAgg}. The results of our ablation experiments are presented in Table \ref{table:ablation}. We use GraphSAGE as our benchmark model for comparison. The explanations about M1, M2 and M3 are as follows:

We conduct an ablation study to verify the impact of each component, using GraphSAGE as the benchmark model. Three components evaluated are as follows:

\begin{itemize}
    \item M1 indicates prototype extracting, feature optimization and GCD attention machenism, mentioned in Sections \ref{chap:met:extfeat}--\ref{chap:met:utlGC}. 
    \item M2 indicates the self-feature matrix, mentioned in Eq. \ref{eq:selfMatrixGeneration}. 
    \item M3 indicates aggregation from typical and atypical perspectives, mentioned in Section \ref{chap:met:GCAgg}.
\end{itemize}

The results indicate that GraphSAGE demonstrates poor performance across all metrics, highlighting its limitations in identifying financial fraud patterns. Conversely, our model exhibits significant improvements in all metrics after incorporating feature transformation and GCD attention mechanisms, which are central to our approach. This underscores the pivotal role of GCD in financial fraud detection. The inclusion of M2 and M3 further enhances the performance of our model, elevating it to a higher level.

\subsection{Impact of GCD on Model Message Aggregation}
\label{chap:exp:impGCDAgg}

% \begin{figure*}[htbp]
%     \centering
%     \begin{subfigure}[b]{0.45\linewidth}
%         \includegraphics[width=\linewidth]{GCAnalyseSimpleTfinance.pdf}
%         \caption{T-Finance Distance Analyse}
%         \label{fig:tfinanceAnalyse}
%     \end{subfigure}
%     \hspace{0.05\linewidth}
%     \begin{subfigure}[b]{0.45\linewidth}
%         \includegraphics[width=\linewidth]{GCAnalyseFDCompSimple.pdf}
%         \caption{FDCompCN Distance Analyse}
%         \label{fig:FDAnalyse}
%     \end{subfigure}
%     \caption{Distance analyse on two datasets. the rate of change is plotted on the right axis, while other data are plotted on the left axis.}
%     \label{fig:distanceAnalyse}
% \end{figure*}

\begin{figure}[bhtp]
    \centering
    \includegraphics[width=\linewidth]{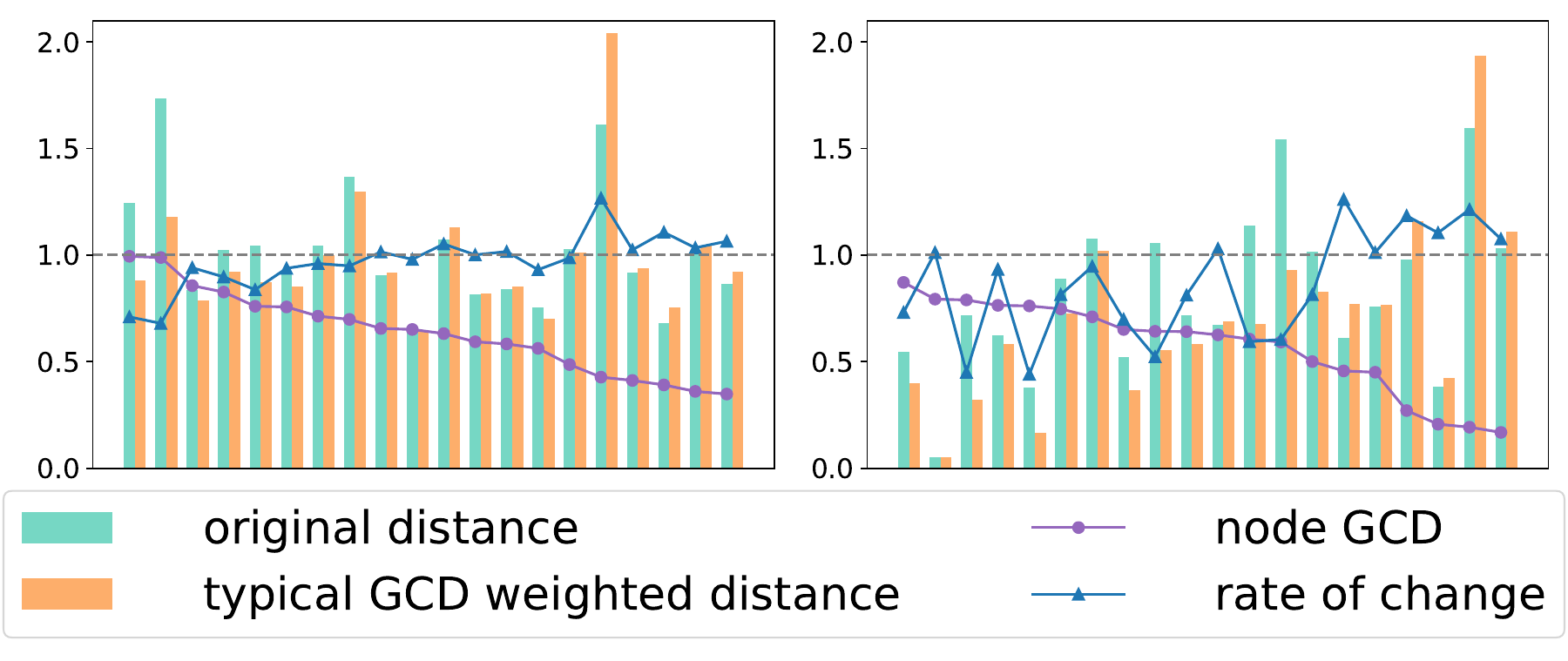}
        \caption{Distances analyze on T-Finance (left) and FDCompCN (right). }
    \label{fig:distanceAnalyse}
\end{figure}

To explore the impact of GCD on model performance and analyze the relationships between nodes and their neighbors from both typical and atypical perspectives. For typical perspective, we examine the typical GCD attentive Euclidean distances  $d_i^{typ}=\frac{\sum_{j\in \mathcal{N}_i}\alpha_{ij}^{typ}\| \mathbf{x}_j- \mathbf{x}_i \|}{\sum_{j\in \mathcal{N}_i}\alpha_{ij}^{typ}}$, where $\alpha_{ij}^{typ}$ is calculated as the method in Section \ref{chap:met:GCAgg}. %as the distance could reflect the similarity of the node and its neighbors. 
For comparision, we also calculate the average Euclidean distances. We randomly choose 20 nodes with neighbors on T-Finance and FDCompCN datasets. The rate of change represents the ratio of the typical GCD-weighted distance to the original distance. The results are reported in Fig. \ref{fig:distanceAnalyse}.
%There are two types of distances considered: first, the original distance, which is the average Euclidean distance between the target node and all its neighbors; second, the GCD-weighted distance, which is the weighted average Euclidean distance between the target node and its neighbors, using GCD values as the weight. The results are presented in Fig. \ref{fig:distanceAnalyse}. The target node GCD value refers to the GCD value of the target node, and the rate of change represents the ratio of the GCD-weighted distance to the original distance.

\begin{figure}[t]
    \centering
    \includegraphics[width=\linewidth]{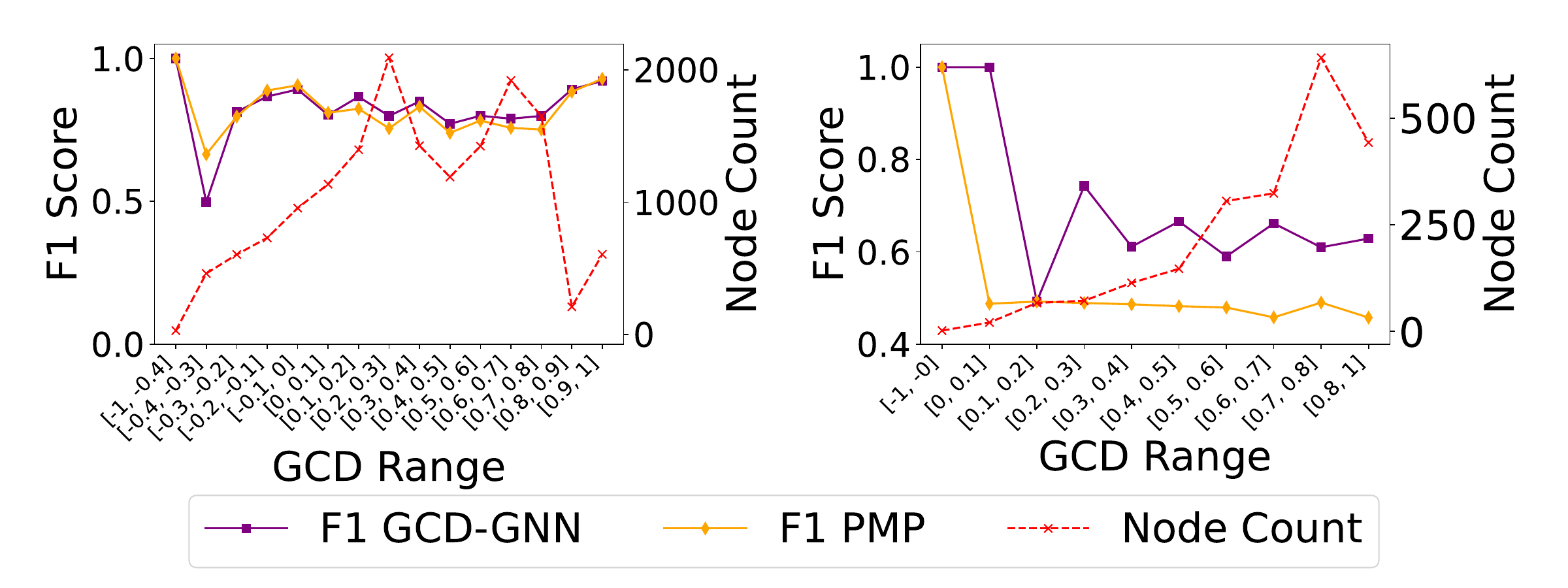}
    \caption{F1-Macro score on GCD-GNN and PMP across different ranges of GCD on T-Finance (left) and FDCompCN (right).}
    \label{fig:f1-tfinance}
\end{figure}

The results show that the rate of change is inversely proportional to the node's GCD. This indicates that higher GCD node tend to aggregate information from closer nodes, while lower GCD tend to aggregate information from more distant nodes. This strategy suggests that nodes with high GCD, which are more typical or representative, tend to aggregate less diverse information, as their characteristics already strongly indicate their belonging to a certain category. Conversely, nodes with lower GCD lack direct distinguishing features and thus tend to rely on diverse information from distant nodes. 
%allowing them to be classified more based on their environment rather than their intrinsic characteristics.
This strategy also ensures that the aggregated information predominantly comes from nodes with higher GCD, making the aggregated information more reliable. 

For atypical perspective, the presence of $g^{atyp}$ allows for the capture of outlier information. We calculate atypical GCD attentive Euclidean
distances $d_i^{atyp}=\frac{\sum_{j\in \mathcal{N}_i}\alpha_{ij}^{atyp}\| \mathbf{x}_j- \mathbf{x}_i \|}{\sum_{j\in \mathcal{N}_i}\alpha_{ij}^{atyp}}$, where $\alpha_{ij}^{atyp}$ is calculated as the method in Section \ref{chap:met:GCAgg}. We find that, $d_i^{atyp}$ tends to be larger compared to $d_i^{typ}$, indicating that extra diverse information can be aggregated from the atypical perspective to aid classification. Detailed results are provided in Appendix.
% 同时该策略聚合的信息均为GC较高的节点的信息，从而让聚合的信息倾向与可信。同时由于$GC_{atyp}$的存在，提供了极端异常信息的捕获。

We analyze the F1-Macro value in the different range of GCD on T-Finance and FDCompCN datasets. The result are reported in Fig. \ref{fig:f1-tfinance} and more metrics %and datasets !!保留则不改
analysis is in Appendix. We compare our model with the most competitive model PMP \cite{zhuo_partitioning_2024}. As shown in Fig. \ref{fig:f1-tfinance}.  We find that GCD-GNN outperforms in most range of GCD, from low to high concretely from 0.1 to 0.8, which demonstrates that: (1) nodes with low GCD absorb more information that differs from their own features, (2) nodes with high GCD absorb more similar features, and (3) incorporating atypical information positively impacts model performance.

\begin{figure}[t]
\centering
\includegraphics[width=0.9\linewidth]{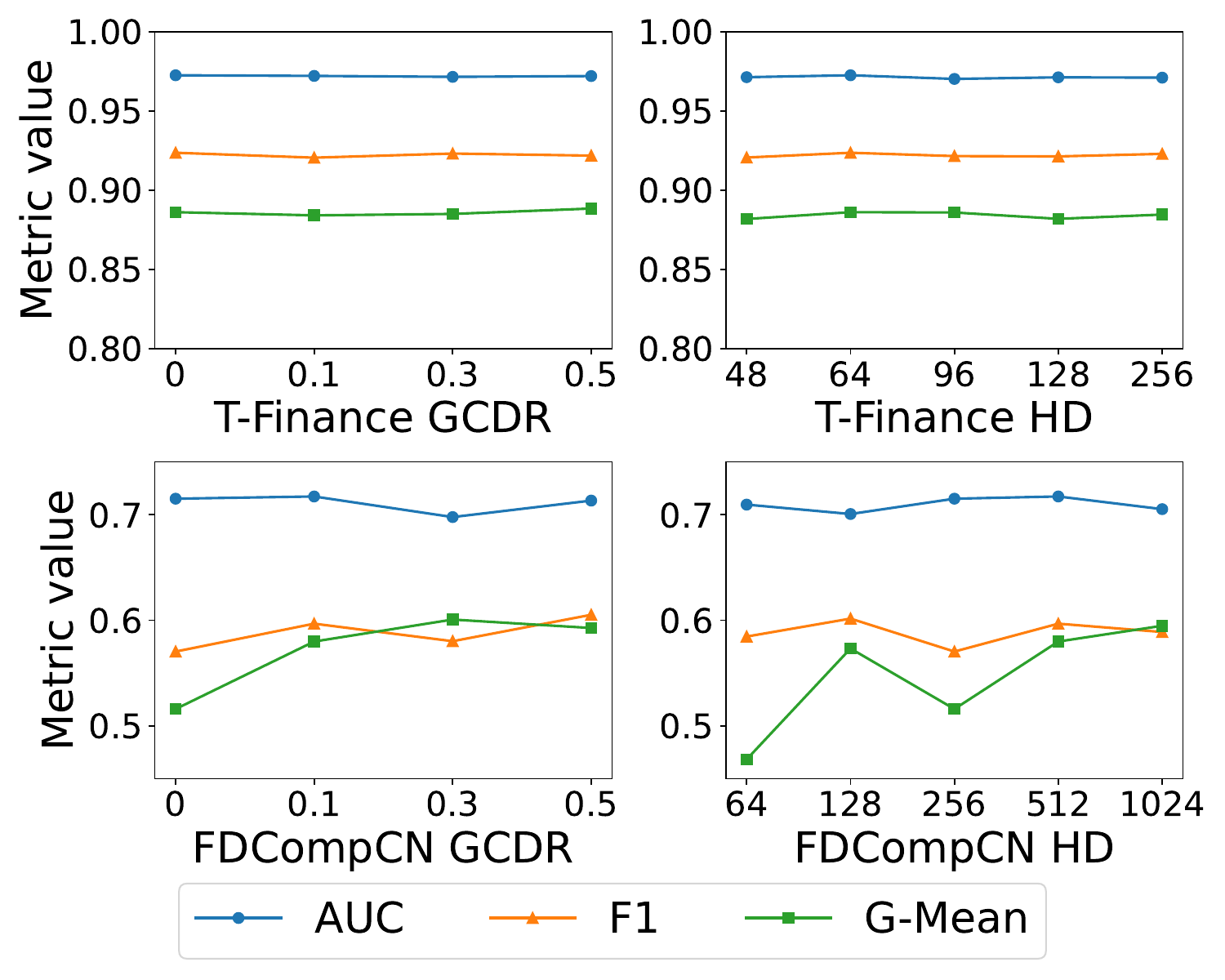}
\raggedright
\caption{Hyperparameters sensitive results.}
%Upper line: T-Finance; Lower line: FDCompCN.}

\label{fig:sensitive}

\end{figure}

\subsection{Sensitive Analyze}
We explore the model’s sensitivity to the important parameters GCD drop rate and hidden dimenssion. All results are presented in Fig. \ref{fig:sensitive}, where GCDR means GCD drop rate and HD means hidden dimension. Detailed values in the figure are provided in the Appendix.

\begin{itemize}
    \item \textbf{The GCD Drop Rate.} During the training of our model, we observed potential overfitting when generating weights through GCD attention mechanism. To address this, in addition to the dropout layer at the end of the network, a GCD dropout layer was incorporated into the model. As a result, we find that the optimal GCD drop rate with the highest AUC is 0 for T-Finance and 0.1 for FDCompCN, suggesting that T-Finance avoids overfitting during the generation of weights, whereas FDCompCN may suffer from slight overfitting. 

    \item \textbf{The Hidden Dimenssion.} The hidden dimension in the model is also crucial to performance; A low hidden dimension leads to inadequate explanation of data features, while a high hidden dimension can result in overfitting. We find that  the model performs best with a feature dimension of 48 on T-Finance, while FDCompCN achieves optimal performance with a dimension of 512, which is proportional to the feature dimension of the respective datasets.
\end{itemize}

% \begin{table*}[hbtp]
%     \centering
%     \caption{Comparison of Methods on Tfinance and FDCompCN}
    
% \end{table*}

% \subsection{Different Fraud Method Task Differences}

\section{Conclusion}
% This paper introduces the Global and Local Confidence Graph Neural Network (GCD-GNN), a novel approach to graph-based anomaly detection designed to address the challenges of heterophily and camouflage in fraudulent activities. By encapsulating the global features of a graph through the introduction of a prototype and calculating a Global Confidence (GCD) value for each node, GCD-GNN effectively distinguishes between benign and fraudulent nodes. The model leverages GCD to generate attention values for message aggregation, enhancing its ability to capture both homophily and heterophily. Extensive experiments on four open datasets demonstrate that GCD-GNN outperforms state-of-the-art models in terms of accuracy and convergence speed while maintaining a compact model size and an expedited training process. Additionally, this paper explores the differences between various fraud detection methods and tasks and presents experimental results that showcase the performance advantages of GCD-GNN in handling nodes with low GCD values, indicating that GCD information can enhance the aggregation process and lead to more accurate classifications of nodes with lower GCD values.

% In summary, the GCD-GNN model provides a new perspective and solution for the field of fraud detection, and its outstanding performance on multiple datasets proves its potential for practical application.

In this work, we introduce the concept of GCD and define its role in the process of information aggregation. We analyze the reasoning behind the effectiveness of GCD in enhancing the detection of fraudulent activities and propose a novel GNN-based model named GCD-GNN. Specifically, our model utilizes GCD for feature optimization, message filtering and aggregation from typical and atypical perspectives. Experimental results demonstrate that GCD-GNN outperforms state-of-the-art methods in terms of AUC, F1-Macro, G-Mean, and convergence speed. We also design a lightweight GCD-GNN (GCD-GNN$_{light}$) that outperforms the baselines on almost all metrics, is slightly weaker than GCD-GNN on fraud detection, but obviously outperforms it in convergence and inference speed.

\bibliography{reference}

\clearpage
% \onecolumn

\appendix

\section{Implementation Details}
The proposed GCD-GNN provides an implementation in PyTorch. All experiments are run on a server with 32 cores, 120GB memory, 1 NVIDIA RTX 4090 GPU, and Ubuntu 20.04 as the operating system. The hyper-parameter setting of GCD-GNN is listed in Table \ref{tab:hyperparameters}. We use grid search to find the best hyperparameters, with results rounded to three decimal places. Detailed results can be found in the configuration files in the config directory within the code. The code is publicly available on Github\footnote{https://github.com/GCDGNN/GCD-GNN/}.
%\footnote{We will make our code publicly available as soon as this article is accepted for publication.}.

\begin{table}[htbp]
    \centering
    \caption{Hyper-parameters setting on T-Finance and FDCompCN datasets.}
    \begin{tabular}{lcc}
        \toprule
        Parameter & T-Finance & FDCompCN \\
        \midrule
        
        learning rate & 0.005 & 0.005\\
        batch size & 1024 &  128 \\
        dropout & 0.292 &  0 \\
        hidden dimension & 64  & 512 \\
        n layer & 1 & 1 \\
        weight decay& 0 & 0 \\
        optimizer & Adam & Adam \\
        thres & 0.5 & 0.5 \\
        GCD drop & 0 & 0.1 \\
         
        \bottomrule
    \end{tabular}
    \label{tab:hyperparameters}
\end{table}

\section{Metrics}
Following \cite{tang2022rethinking}, we use AUC, F1-Macro and G-Mean as our experiments metrics. AUC measures the area under the ROC curve and reflects the model’s ability to distinguish between positive and negative classes across all possible classification thresholds. F1-Macro calculates the F1 score for each class independently and then takes the average. The G-Mean, or geometric mean, is the square root of the product of sensitivity and specificity, showing the balance between true positive rate and true negative rate. Higher values for these metrics indicate better method performance.

% \section{Code Address}

\section{Baseline Models Introduction}

In this section, we describe the baseline models used for comparison.

The general models are as follows:

\begin{itemize}
    \item GCN \cite{kipf2016semi}, A graph convolutional network utilizing the first-order approximation of localized spectral filters on graphs.
    \item GAT \cite{velivckovic2017graph}, A graph attention network that employs the attention mechanism for neighbor aggregation.
    \item GraphSAGE \cite{hamilton2017inductive}, A graph neural network model based on sampling a fixed number of neighbor nodes.
\end{itemize}
The fraud detection models are as follows:

\begin{itemize}
    \item Care-GNN \cite{dou2020enhancing}, A camouflage-resistant GNN that enhances the aggregation process with three unique modules designed to counter camouflages and incorporates reinforcement learning.
    \item PC-GNN \cite{liu2021pick}, A GNN-based method for addressing category imbalance in graph-based fraud detection through resampling techniques.
    \item BWGNN \cite{tang2022rethinking}, A graph neural network utilizing a label-aware high-frequency indicator to prune the heterogeneous edges, effectively reducing heterophily and boosting graph anomaly detection performance.
    \item SplitGNN \cite{wu2023splitgnn}, A spectral GNN that addresses fraud detection in heterophilic graphs by splitting the graph into subgraphs and applying band-pass filters to capture diverse frequency signals.
    \item GHRN \cite{gao2023addressing}, A graph neural network using Beta wavelet filters to improve anomaly detection by addressing spectral energy 'right-shift' in large-scale datasets.
    \item PMP \cite{zhuo_partitioning_2024}, A graph neural network enhancing fraud detection by distinguishing between homophilic and heterophilic neighbors in message passing, addressing label imbalance and mixed homophily-heterophily.
\end{itemize}

\section{Training AUC and Time Details}

In Tabel \ref{tab:time_performance} we present the detailed AUC value and convergence time consumption.

\begin{table}[htbp]
\centering
\caption{Traing AUC and time.}
\begin{tabular}{ccc}
\toprule
\textbf{model} & \textbf{AUC} & \textbf{Time (s)} \\
\midrule

PCGNN & 93.49 & 369.40 \\
Care-GNN & 87.22 & 287.09 \\
BWGNN & 92.33 & 16.04 \\
SplitGNN & 95.51 & 5592.14 \\
GHRN & 95.78 & 191.42\\
PMP & 97.07 & 1661.78 \\
GCD-GNN$_{light}$ & 97.06 & 97.68 \\
GCD-GNN & 97.26 & 624.38 \\
\bottomrule
\end{tabular}
\label{tab:time_performance}
\end{table}

\section{Sensitive Analyze Details}
In Tables \ref{tab:tfinance_performance_hid_dim}--\ref{tab:fdcompcn_performance_drop_rate}, We present the detailed value of AUC, F1, G-Mean influenced by hyperparameters.
\begin{table}[htbp]
\centering
\caption{Performance metrics for different hidden dimension on T-Finance.}
\begin{tabular}{lccc}
\toprule
\textbf{hiddim} & \textbf{AUC} & \textbf{F1-Macro} & \textbf{G-Mean} \\
\midrule
48  & 97.14$_{\scriptsize{\pm 0.01}}$ & 92.07$_{\scriptsize{\pm 0.03}}$ & 88.19$_{\scriptsize{\pm 0.10}}$  \\
64  & 97.26$_{\scriptsize{\pm 0.01}}$ & 92.37$_{\scriptsize{\pm 0.05}}$ & 88.62$_{\scriptsize{\pm 0.11}}$  \\
96  & 97.03$_{\scriptsize{\pm 0.02}}$ & 92.16$_{\scriptsize{\pm 0.02}}$ & 88.60$_{\scriptsize{\pm 0.07}}$  \\
128 & 97.13$_{\scriptsize{\pm 0.01}}$ & 92.14$_{\scriptsize{\pm 0.03}}$ & 88.20$_{\scriptsize{\pm 0.12}}$  \\
256 & 97.11$_{\scriptsize{\pm 0.02}}$ & 92.30$_{\scriptsize{\pm 0.01}}$ & 88.47$_{\scriptsize{\pm 0.06}}$  \\
\bottomrule
\end{tabular}
\label{tab:tfinance_performance_hid_dim}
\end{table}

\begin{table}[htbp]
\caption{Performance metrics for FDCompCN with different hidden dimensions.}
\centering
\begin{tabular}{lccc}
\toprule
\textbf{hiddim} & \textbf{AUC} & \textbf{F1-Macro} & \textbf{G-Mean} \\
\midrule
64   & 70.95$_{\scriptsize{\pm 0.14}}$ & 58.46$_{\scriptsize{\pm 0.61}}$ & 46.83$_{\scriptsize{\pm 2.25}}$ \\
128  & 70.06$_{\scriptsize{\pm 0.12}}$ & 60.16$_{\scriptsize{\pm 0.21}}$ & 57.33$_{\scriptsize{\pm 0.40}}$ \\
256  & 71.51$_{\scriptsize{\pm 0.12}}$ & 57.04$_{\scriptsize{\pm 0.53}}$ & 51.60$_{\scriptsize{\pm 2.51}}$ \\
512  & 71.72$_{\scriptsize{\pm 0.18}}$ & 59.68$_{\scriptsize{\pm 0.31}}$ & 57.99$_{\scriptsize{\pm 0.31}}$ \\
1024 & 70.52$_{\scriptsize{\pm 0.10}}$ & 58.89$_{\scriptsize{\pm 0.18}}$ & 59.48$_{\scriptsize{\pm 0.20}}$ \\
\bottomrule
\end{tabular}
\label{tab:fdcompcn_performance_hid_dim}
\end{table}

\begin{table}[htbp]
\caption{Performance metrics on T-Finance with different attention drop rates.}
\centering
\begin{tabular}{lccc}
\toprule

\textbf{GCD\_drop} & \textbf{AUC} & \textbf{F1-Macro} & \textbf{G-Mean} \\

\midrule
0   & 97.26$_{\scriptsize{\pm 0.01}}$ & 92.37$_{\scriptsize{\pm 0.05}}$ & 88.62$_{\scriptsize{\pm 0.11}}$ \\
0.1 & 97.22$_{\scriptsize{\pm 0.02}}$ & 92.06$_{\scriptsize{\pm 0.02}}$ & 88.42$_{\scriptsize{\pm 0.08}}$ \\
0.3 & 97.16$_{\scriptsize{\pm 0.02}}$ & 92.32$_{\scriptsize{\pm 0.02}}$ & 88.51$_{\scriptsize{\pm 0.07}}$ \\
0.5 & 97.21$_{\scriptsize{\pm 0.02}}$ & 92.19$_{\scriptsize{\pm 0.02}}$ & 88.85$_{\scriptsize{\pm 0.15}}$ \\
\bottomrule
\end{tabular}
\label{tab:tfinance_performance_drop_rate}
\end{table}

\begin{table}[t!]
\caption{Performance metrics for FDCompCN with different attention drop rates.}
\centering
\begin{tabular}{lccc}
\toprule
\textbf{GCD\_drop} & \textbf{AUC} & \textbf{F1-Macro} & \textbf{G-Mean} \\
\midrule
0   & 71.51$_{\scriptsize{\pm 0.12}}$ & 57.04$_{\scriptsize{\pm 0.53}}$ & 51.60$_{\scriptsize{\pm 2.51}}$ \\
0.1 & 71.72$_{\scriptsize{\pm 0.18}}$ & 59.68$_{\scriptsize{\pm 0.31}}$ & 57.99$_{\scriptsize{\pm 0.31}}$ \\
0.3 & 69.77$_{\scriptsize{\pm 0.20}}$ & 58.02$_{\scriptsize{\pm 0.43}}$ & 60.07$_{\scriptsize{\pm 0.25}}$ \\
0.5 & 71.33$_{\scriptsize{\pm 0.04}}$ & 60.53$_{\scriptsize{\pm 0.06}}$ & 59.27$_{\scriptsize{\pm 0.35}}$ \\
\bottomrule
\end{tabular}
\label{tab:fdcompcn_performance_drop_rate}
\end{table}

\section{Performance in the Different Range of GCD on T-Finance and FDCompCN Datasets}
\begin{figure}[t!]
    \centering
    \begin{subfigure}[b]{\linewidth}
        \centering
        \includegraphics[width=\linewidth]{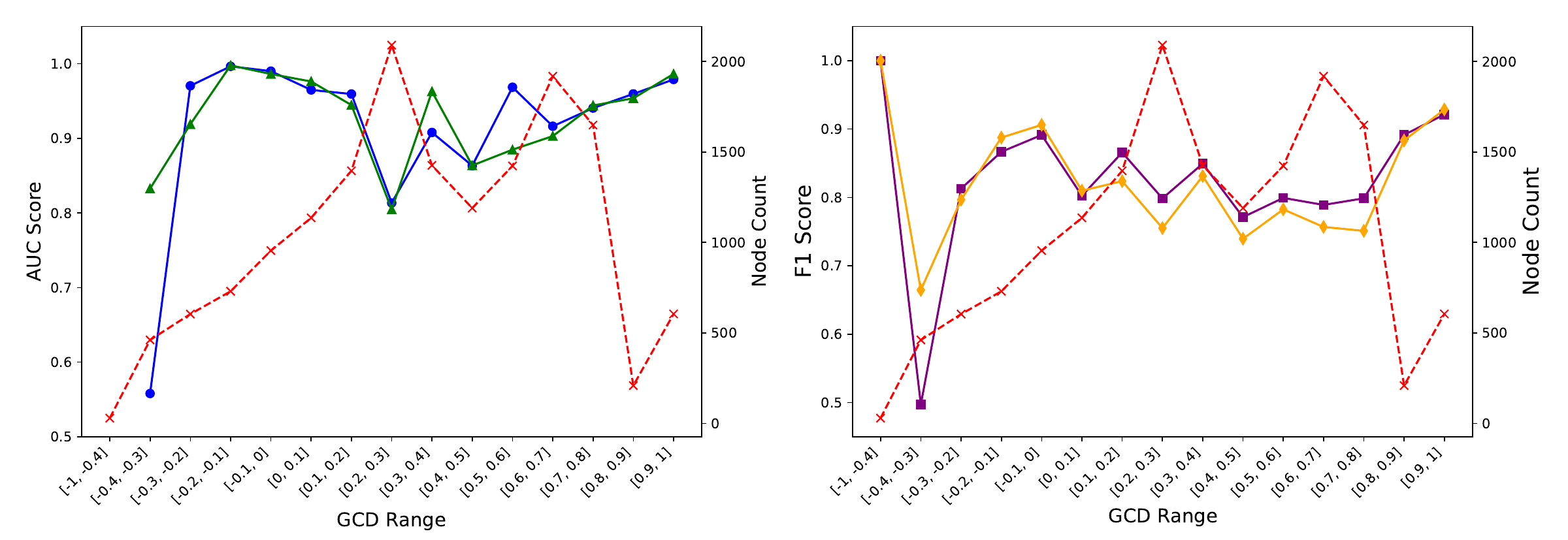}
    \end{subfigure}
    \begin{subfigure}[b]{0.98\linewidth}
        \centering
        \includegraphics[width=\linewidth]{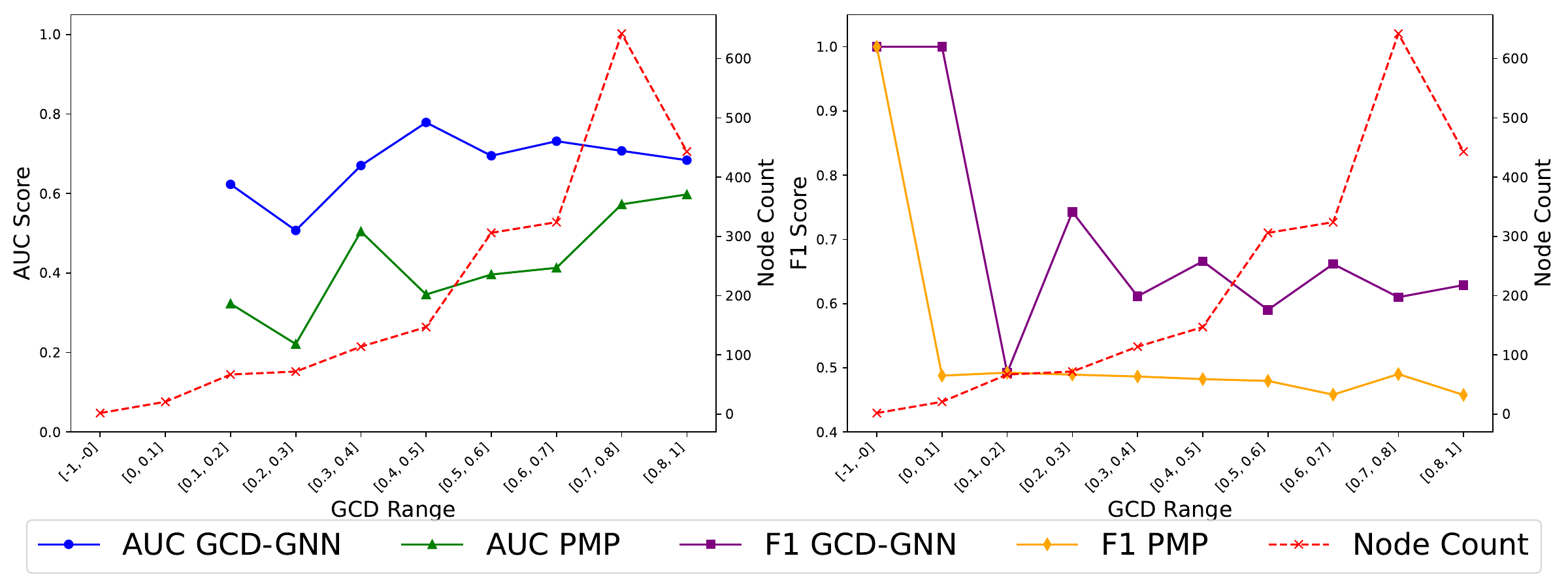}
    \end{subfigure}
    \caption{Performance in the different range of GCD on T-Finance (top) and FDCompCN (bottom) datasets}
    \label{fig:f1auc_gc}
\end{figure}

We visualize AUC and F1-MARCO in diffrent range on the test set on T-Finance and FDCompCN datasets, as shown in Fig. \ref{fig:f1auc_gc}. The missing AUC values are due to the presence of only one category of nodes within the specific GCD range.

\section{Typical and Atypical GCD weighted Distance Analysis}
We calculate the atypical GCD weighted distance according to Section \ref{chap:exp:impGCDAgg}. Typical and atypical GCD weighted distances are calculated as follows:

\begin{align}
    d_i^{typ}=\frac{\sum_{j\in \mathcal{N}_i}\alpha_{ij}^{typ}\| \mathbf{x}_j- \mathbf{x}_i \|}{\sum_{j\in \mathcal{N}_i}\alpha_{ij}^{typ}}, \notag \\
    d_i^{atyp}=\frac{\sum_{j\in \mathcal{N}_i}\alpha_{ij}^{atyp}\| \mathbf{x}_j- \mathbf{x}_i \|}{\sum_{j\in \mathcal{N}_i}\alpha_{ij}^{atyp}}, 
\end{align}
where $\alpha_{ij}^{typ}$ and $\alpha_{ij}^{atyp}$ are calculated as the method mentioned in Section \ref{chap:met:GCAgg}.

\begin{figure}[t!]
    \centering
    \includegraphics[width=\linewidth]{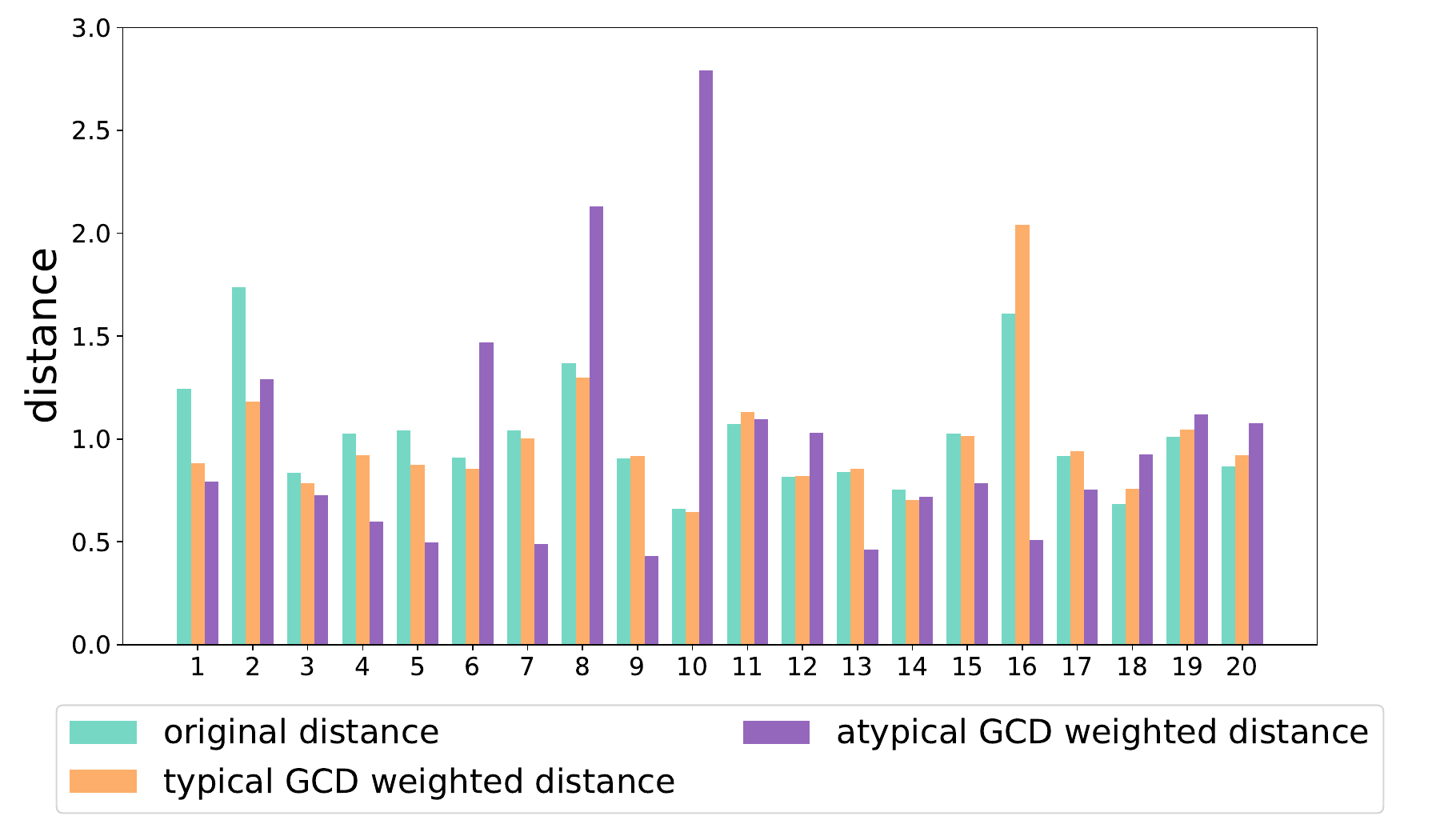}
    \caption{GCD weighted distance analysis}
    \label{fig:atyDist}
\end{figure}

As the result shown in Fig. \ref{fig:atyDist}, we find that, $d_i^{atyp}$ tends to be larger compared to $d_i^{typ}$, indicating that extra diverse information can be aggregated from the atypical perspective to aid classification.

% \section{Detailed Datasets Description}

% \begin{table*}[htbp]
%     \centering
%     \caption{Statistics on datasets}
%     \begin{tabular}{l|cccccc}
%         \toprule
%         & \multicolumn{4}{c}{Statistics} \\
%         Datasets & Nodes  & Anomaly(\%) & Features & Relations & Edges \\
%         \midrule
%         T-Finance & 39,357 & 4.58\% & 10 &  homo&21,222,543\\
%         \midrule
%         \multirow{4}{*}{FDCompCN} & \multirow{4}{*}{5,317} & \multirow{4}{*}{10.5\%} & \multirow{4}{*}{57}  & C-I-C & 5,686 \\ 
%                           &       &       &    & C-P-C & 760   \\ 
%                           &       &       &    & C-S-C & 1,043  \\ 
%                           &       &       &     & Homo  & 7,407  \\ 
        
%         \bottomrule
%     \end{tabular}
%     \label{table:statdataset}
% \end{table*}

\end{document}